\documentclass[11pt]{article}
\usepackage{acl}
\usepackage{times}
\usepackage{latexsym}
\usepackage[T1]{fontenc}
\usepackage[utf8]{inputenc}
\usepackage{microtype}
\usepackage{inconsolata}
\usepackage{graphicx}
\usepackage{amsmath}
\usepackage{booktabs}
\usepackage{tabularx}
\usepackage{fontawesome5}
\usepackage{pgfplots}
\pgfplotsset{compat=1.18}
\usetikzlibrary{patterns}

\title{RefactorPlatform: An Open-Source Harness for Controlled Evaluation of Repository-Scale Refactoring Agents}

\author{
 \textbf{Aziz Ben Amor\textsuperscript{1,*}},
 \textbf{Drish Mali\textsuperscript{1,*}},
 \textbf{Mann Acharya\textsuperscript{1,*}},
 \\
 \textbf{Vijayasri Iyer\textsuperscript{1,\textdagger}},
 \textbf{Sébastien Bratières\textsuperscript{1, 2}}
\\
\\
 \textsuperscript{1}Pi School,
 \textsuperscript{2}Translated
}

\begin{document}
\maketitle

\begin{abstract}
Repository-scale refactoring requires coding agents to propagate a single change across many interdependent files without altering program behavior, yet to our knowledge no existing harness isolates the design choices that determine agent success on this task. We present \textbf{RefactorPlatform}, an open-source evaluation harness that holds the environment fixed and varies each design axis explicitly: model backbone (via OpenRouter and GitHub Copilot CLI), execution regime (baseline, retrieval-augmented, and multi-agent), and prompt specificity. Each run executes in an isolated workspace with live terminal streaming, per-task logging of tokens, diffs, and transcripts, AST-based verification, and exportable telemetry for audit and reproduction. Demonstrating the platform on 100 multi-file RefactorBench tasks across four model families, we illustrate the analyses it supports: AST-aware chunking outperforms naive token-window chunking by 25--30 percentage points (pp) across prompt modes, whereas naive retrieval falls below the retrieval-free baseline; a lean retrieval-augmented single agent (86\%) beats multi-agent execution  (66\%) on matched tasks, with no task passing under delegation that fails under retrieval; and retrieval's accuracy gains absorb its token overhead, leaving cost per successful refactoring unchanged. RefactorPlatform is open-sourced to make refactoring-agent evaluation reproducible and auditable.
\end{abstract}

\def\thefootnote{*}\footnotetext{Equal contribution.}
\def\thefootnote{\textdagger}\footnotetext{Corresponding author: \href{mailto:vijayasri.iyer@picampus-school.com}{vijayasri.iyer@picampus-school.com}}
\def\thefootnote{\textdaggerdbl}\footnotetext{\href{https://github.com/PiSchool/refactor-platform}{\faGithub\ github.com/PiSchool/refactor-platform}}
\def\thefootnote{\arabic{footnote}}

\section{Introduction}
Refactoring, ``improving the internal structure of software without altering its external behavior'' \citep{Fowler1999Refactoring}, has shifted from localized IDE transformations to repository-scale campaigns that propagate a single change across many interdependent files, where manual propagation is error-prone and a persistent source of technical debt \citep{FERNANDES2020106347}. Function-level benchmarks reveal little about this regime \citep{Liuetal2023}, and while repository-scale suites like SWE-bench better reflect production engineering \citep{Jimenezetal2023}, refactoring imposes a stricter requirement: edits must be \emph{behavior-preserving} while still propagating across files. RefactorBench targets exactly this setting with AST-verified multi-file tasks \citep{gautam2025refactorbenchevaluatingstatefulreasoning}, and SWE-Refactor \citep{xu2026swerefactorrepositorylevelbenchmarkrealworld} verifies developer-written refactorings through compilation, tests, and RefactoringMiner \citep{tsantalis2022refactoringminer}.

Current agents remain inadequate at this scale \citep{siddeeq2025llm, guan2025repotransagentmultiagentllmframework}. A single agent perceives only a small slice of the repository at any moment, and most Retrieval-Augmented Generation (RAG) pipelines are \emph{example-based}, retrieving historical snippets rather than performing structural discovery over the live code a refactoring must touch \citep{xu2026swerefactorrepositorylevelbenchmarkrealworld}, despite evidence that AST-aware chunking retrieves more coherent units \citep{zhang2025castenhancingcoderetrievalaugmented}. The result is \emph{structural blindness}: systems propose a change but miss sites where it must be applied. A common response is adding more agents in specialized roles \citep{Heetal2024, Qianetal2023, Phanetal2024, Dongetal2025}, yet the evidence is mixed: multi-agent systems lose to single agents on executability and consistency while adding handoffs, tokens, and drift \citep{Yinetal2025, Guoetal2025}. Evidence on repository-scale \emph{refactoring} specifically remains fragmented. Mature harnesses exist for issue resolution: SWE-agent defines agent--computer interfaces \citep{yang2024sweagent}, OpenHands provides an open platform with a pluggable evaluation harness \citep{wang2025openhands}, and Agentless \citep{xia2025agentless} shows that lean pipelines can rival agentic scaffolds. To our knowledge, however, no comparable infrastructure isolates retrieval, prompting, orchestration, and model routing for repository-scale refactoring under controlled, auditable logging.

\begin{figure*}[htbp]
    \centering
    \includegraphics[width=\textwidth]{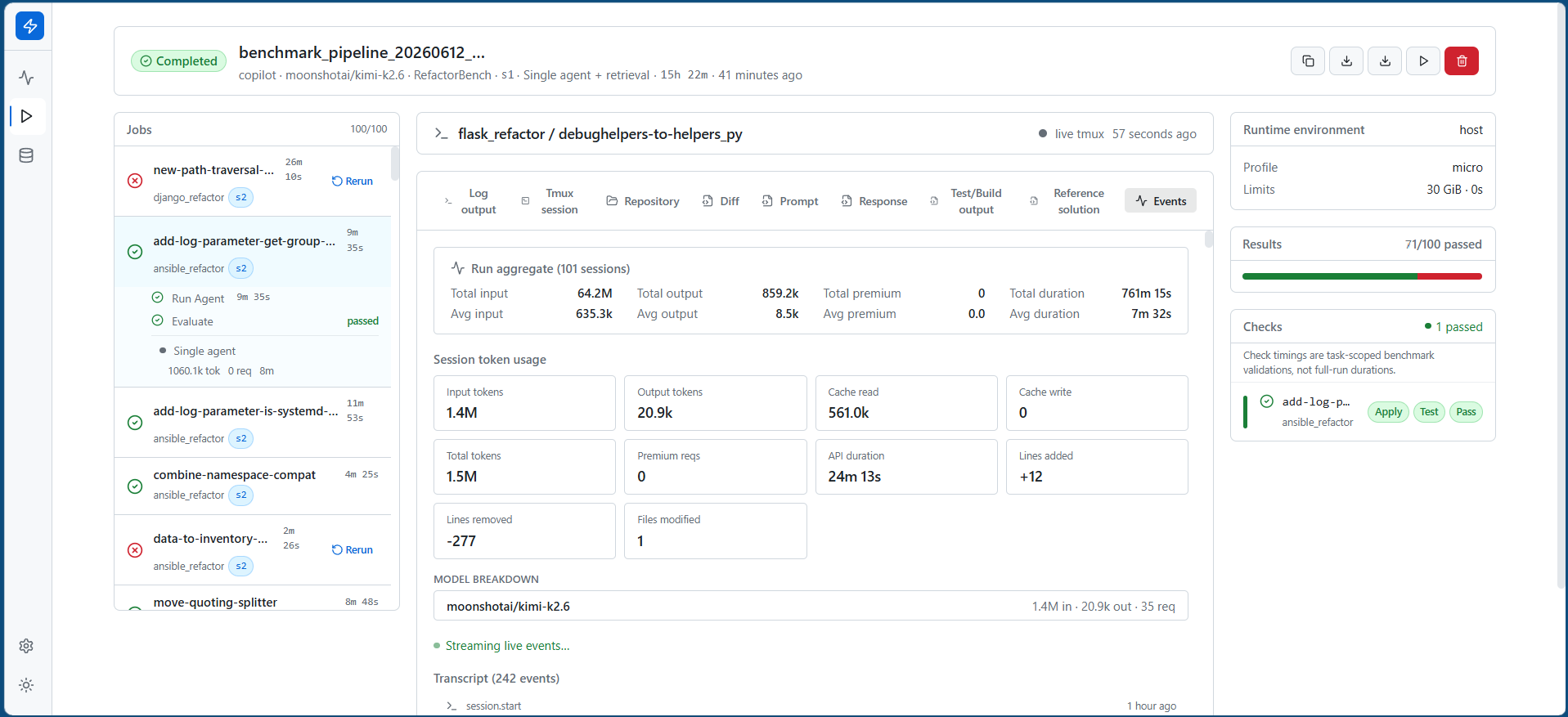}
    \caption{The RefactorPlatform web interface monitoring a completed 100-task run. The dashboard unifies real-time job scheduling, token-consumption analytics, and live terminal streams in a single view.}
    \label{fig:dashboard_screenshot}
\end{figure*}

We close this gap with \textbf{RefactorPlatform}, an open-source evaluation harness that holds the environment fixed and varies each design axis explicitly, across three cumulative regimes (Section~\ref{subsec:execution_regimes}): a baseline terminal assistant ($S_1$), a retrieval-augmented configuration ($S_2$), and an orchestrated multi-agent workflow ($S_3$). The platform targets researchers benchmarking coding agents and practitioners auditing agent configurations before deployment. Figure~\ref{fig:dashboard_screenshot} shows the dashboard view of a completed 100-task campaign. The platform lets an operator answer questions like \textbf{Q1:} how much does AST-aware chunking improve pass rates over baseline execution and token-window chunking? \textbf{Q2:} does multi-agent execution recover what it costs, relative to lean single-agent retrieval? \textbf{Q3:} how far does structural retrieval compensate for
under-specified prompts?

Our contributions are: (i) \textbf{RefactorPlatform} isolates
per-task workspaces, four independently configurable design axes, live
terminal streaming, and exportable per-task telemetry; (ii) an
\textbf{AST verification pipeline}, AST-based unit tests for Python
and extraction--compilation--test--RefactoringMiner gates for Java, so that
reported pass rates reflect behavior-preserving change rather than
plausible-looking diffs; and (iii) a \textbf{100-task demonstration campaign} across four model families illustrating the analyses the platform enables: a prompt-sensitivity map, a context-layer comparison in which AST-aware retrieval dominates LSP diagnostics, and a best-case pass rate of 86\% at \$0.13 per success.

\section{RefactorPlatform Architecture and Workflow}
RefactorPlatform serves as an evaluation harness designed specifically to benchmark and audit large language model coding agents executing repository-scale refactoring tasks. The platform addresses the lack of standardized, isolated testing infrastructure required for multi-file source code modifications. Each task runs in an isolated workspace to prevent state leakage between runs. During each turn of an agent session, the harness captures all generated file diffs, full terminal execution logs, precise input and output token consumption, and complete interaction transcripts streaming through persistent Tmux sessions.
 
Operators configure and execute evaluations via the platform's web interface across four explicit design variables:
\begin{itemize}
    \item \textbf{Model Selection:} Supports any model available through OpenRouter (BYOK) or the GitHub Copilot CLI \citep{githubcopilotcli}.
    \item \textbf{Execution Regimes:} three cumulative regimes including an $S_1$ baseline terminal assistant, an $S_2$ task-scoped structural retrieval stack utilizing the Model Context Protocol (MCP), and an $S_3$ regime that enables the agent's native sub-agent agents.
    \item \textbf{Prompt Modes:} The architecture sweeps across Descriptive, Base, and Lazy specificity levels to systematically map how agents adapt to under-specified or ambiguous human requests.
    \item \textbf{Target Dataset Tasks:} Evaluation targets are drawn from integrated benchmark suites.
\end{itemize}

\subsection{Natural Language Prompt Modes}
To measure agent resilience to varying degrees of instruction clarity, each task is
evaluated under one of three prompt configurations, following the specificity levels
defined by RefactorBench \citep{gautam2025refactorbenchevaluatingstatefulreasoning}.
\textit{Descriptive} mode supplies full guidance on what to change, where to locate the
relevant elements, and how to perform the refactoring (What + Where + How).
\textit{Base} mode provides minimal instructions specifying only the structural
refactoring type and the target entity (What + Where).
\textit{Lazy} mode simulates highly ambiguous real-world usage, supplying only a vague
refactoring type with no positional or implementation detail (What only).  

\subsection{Execution Regimes}
\label{subsec:execution_regimes}
We implement three execution regimes to separate the contribution of terminal tooling, retrieval, and sub-agent delegation, along with the cost each carries. This helps us to isolate whether performance gains stem from the raw capabilities of the model backbone, codebase indexing extensions, or complex multi-agent layouts, along with the cost trade-offs associated with each.

\paragraph{Regime $S_1$ (Baseline Assistant):}
The single agent operates within a standalone terminal CLI session. It has native access to fundamental localized workspace utilities, including filesystem parsing, directory listing, regex tools, and direct execution shells (\texttt{grep}, \texttt{view}, \texttt{edit}, \texttt{bash}), but lacks external context access. 

\paragraph{Regime $S_2$ (Retrieval-Augmented Assistant):}
This regime extends the $S_1$ framework by integrating a persistent, task-scoped codebase search engine via MCP. Prior to execution, repositories are chunked  by CocoIndex \citep{cocoindex} and converted into vector embeddings using \texttt{nomic-embed-code} encoder (GGUF Q4\_K\_M optimization).

The $S_2$ agent has codebase discovery capabilities, including \texttt{search\_codebase}, \texttt{list\_indexed\_files}, and \texttt{read\_indexed\_file}. Queries are resolved using a hybrid vector similarity (cosine metric) and lexical keyword matching architecture (BM25) \cite{robertson2009bm25}, with results fused via Reciprocal Rank Fusion (RRF) \cite{cormack2009rrf}.



\paragraph{Regime $S_3$ (Sub-Agent Delegation):}
$S_3$ gives the agent its native sub-agent capability and a prompt block
describing when to delegate. We impose no topology. The agent decides whether
to decompose the task and how, which means the regime is only exercised when
it chooses to use it. So the harness proves delegation from the agent's own
event stream and marks each run \texttt{conformant} or
\texttt{setup\_not\_exercised} regardless of whether the task passed. A
correct edit made without delegating is not evidence about delegation.
The platform also supports an external orchestrator, the CLI Agent
Orchestrator \citep{awslabs2026cao}, with fixed supervisor, analyst, modifier,
and validator roles. That is a separate configuration and not the system in
Table~\ref{tab:main_results}.

\subsection{Benchmarks}

We evaluate our platform using \textbf{RefactorBench}, a Python benchmark designed to measure complex navigational reasoning across real-world codebases. This benchmark consists of 100 multi-file tasks across nine repositories (e.g., Flask, FastAPI). It requires agents to modify between 2 and 31 files per task, moving beyond simple code generation to "stateful reasoning." Success is verified through AST-based unit tests that ensure structural correctness \citep{gautam2025refactorbenchevaluatingstatefulreasoning}. The platform additionally integrates the Java-centric \textbf{SWE-Refactor} suite \citep{xu2026swerefactorrepositorylevelbenchmarkrealworld}. Preliminary results on it are reported in Appendix~\ref{sec:swe-refactor}.

\subsection{User Interface and Dashboard Workflow}
\label{subsec:dashboard_workflow}
An evaluation run is configured and monitored through the RefactorPlatform web interface, which provides a centralized dashboard for real-time monitoring of long-horizon tasks (Figure~\ref{fig:dashboard_screenshot}). The interface is structured into three primary columns:
\begin{itemize}
    \item \textbf{Task Queue (Left):} Tracks real-time execution status, runtimes, and success badges for all benchmark jobs, featuring one-click execution triggers for isolated retries.
    \item \textbf{Workspace \& Execution Traces (Center):} Aggregates live performance metrics such as token consumption, API request counts, and line diffs, alongside tabbed views for persistent \texttt{tmux} terminal streams, repository file trees, prompt mode configs, and verbose test outputs.
    \item \textbf{Environment Control Panel (Right):} Displays host container resource allocations (e.g., memory limits) and tracks cumulative success metrics as tasks complete.
\end{itemize}

To launch an evaluation, the operator selects a target benchmark: either the Python navigation tasks in RefactorBench or the Java compilation tasks in SWE-Refactor with the defined configuration parameters across the four design axes. Asynchronous workers then deploy the designated agents into isolated workspace checkouts subject to operational timeouts, streaming telemetry directly back to the dashboard. 



\subsection{AST Verification Pipeline}
Once a run finishes, every generated patch has to clear a conjunctive verification gate before it counts as correct. For Python, that means passing all associated AST-based unit tests. Java is held to a higher bar: code extraction, successful compilation, and a full test pass, all at once, plus a structural check via \textit{RefactoringMiner} to weed out lazy, non-functional stubs that technically compile but don't actually refactor anything. CodeBLEU \citep{Ren2020CodeBLEUAM} scores are also logged, though only as a background diagnostic - they don't factor into pass/fail. When a run is done, the operator can review it live and export the full telemetry, diagnostics, and transcripts as a ZIP or CSV archive.

\subsection{Performance Metrics} We evaluate our experimental setups using two primary criteria:\begin{itemize}

\item \textbf{Task Pass Rate ($\mathit{PR}_{\text{task}}$)}: The mean success rate across $N$ tasks, where a task is successful ($I_j = 1$) only if it passes all $K_j$ associated AST unit tests ($\text{ASTTest}_{j,k} \in \{0, 1\}$), allowing no partial credit:\begin{equation}
\mathit{PR}_{\text{task}} = \frac{1}{N} \sum_{j=1}^{N} I_j, \quad I_j = \bigwedge_{k=1}^{K_j} \mathit{ASTTest}_{j,k}
\end{equation}

\item \textbf{Cost-efficiency ($CE$)}: The average financial token cost ($C_j$) incurred per successful task, penalizing inefficient token consumption: \begin{equation}CE = \sum_{j=1}^{N} C_j / \sum_{j=1}^{N} I_j.  \end{equation}
\end{itemize}

\section{Results and Empirical Analysis}
We evaluate on the 100 multi-file Python tasks in RefactorBench.
Because full factorial sweeps across frontier models are cost-prohibitive,
we run a two-stage \textbf{demonstration campaign} under single-run budgets:
a full ablation on \texttt{qwen3.6-flash} (Table~\ref{tab:main_results}),
then cross-backbone $S_1$ vs.\ full $S_2$ pipeline comparisons
(Table~\ref{tab:final_results}).
These outcomes illustrate the analyses operators can obtain from the harness
at modest cost; the same configuration surface supports larger campaigns
when additional compute is available.

\begin{table*}[h]
\centering
\small
\caption{Ablation on the \texttt{qwen3.6-flash}: prompt specificity, LSP feedback, and chunking strategy.}
\begin{tabular}{lllll}
\toprule
\textbf{Model} & \textbf{Regime} & \textbf{Prompt Mode} & \textbf{Augmentation} & \textbf{Pass Rate} \\
\midrule
\texttt{qwen3.6-flash} & $S_1$ Baseline & Descriptive & +LSP Enforcement & 73\% \\
\texttt{qwen3.6-flash} & $S_1$ Baseline & Descriptive & -LSP Enforcement & 71\% \\
\texttt{qwen3.6-flash} & $S_1$ Baseline & Base & +LSP Enforcement & 64\% \\
\texttt{qwen3.6-flash} & $S_1$ Baseline & Base & -LSP Enforcement & 59\% \\
\texttt{qwen3.6-flash} & $S_1$ Baseline & Lazy & +LSP Enforcement & 48\% \\
\texttt{qwen3.6-flash} & $S_1$ Baseline & Lazy & -LSP Enforcement & 44\% \\
\midrule
\texttt{qwen3.6-flash} & $S_2$ Retrieval & Descriptive & AST Chunking & \textbf{86\%} \\
\texttt{qwen3.6-flash} & $S_2$ Retrieval & Base & AST Chunking & 74\% \\
\texttt{qwen3.6-flash} & $S_2$ Retrieval & Lazy & AST Chunking & 56\% \\
\texttt{qwen3.6-flash} & $S_2$ Retrieval & Descriptive & Naive Chunking & 57\% \\
\texttt{qwen3.6-flash} & $S_2$ Retrieval & Base & Naive Chunking & 44\% \\
\texttt{qwen3.6-flash} & $S_2$ Retrieval & Lazy & Naive Chunking & 31\% \\
\midrule
\texttt{qwen3.6-flash} & $S_3$ Orchestrated & Descriptive & Copilot Sub-Agents & 66\% \\
\bottomrule

\end{tabular}

\label{tab:main_results}
\end{table*}

\begin{table*}[h]
\centering
\small
\caption{Cross-model results evaluating structural retrieval impact across four distinct model architectures comparing $S_1$ Baseline vs $S_2$ AST Chunking based Retrieval with descriptive prompts}
\begin{tabular}{lllll}
\toprule
\textbf{Model} & \textbf{Regime} & \textbf{Prompt Mode} & \textbf{Augmentation} & \textbf{Pass Rate} \\
\midrule
\texttt{qwen3.6-flash} & $S_1$ Baseline & Descriptive & \multicolumn{1}{c}{-} & 73\% \\
\texttt{qwen3.6-flash} & $S_2$ Retrieval & Descriptive & AST Chunking & 86\% \\
\midrule
\texttt{minimax-m3} & $S_1$ Baseline & Descriptive & \multicolumn{1}{c}{-} & 75\% \\
\texttt{minimax-m3} & $S_2$ Retrieval & Descriptive & AST Chunking & 81\% \\
\midrule
\texttt{kimi-k2.6} & $S_1$ Baseline & Descriptive & \multicolumn{1}{c}{-} & 71\% \\
\texttt{kimi-k2.6} & $S_2$ Retrieval & Descriptive & AST Chunking & 78\% \\
\midrule
\texttt{deepseek-v4-pro} & $S_1$ Baseline & Descriptive & \multicolumn{1}{c}{-} & 77\% \\
\texttt{deepseek-v4-pro} & $S_2$ Retrieval & Descriptive & AST Chunking & \textbf{89\%} \\
\bottomrule
\end{tabular}

\label{tab:final_results}
\end{table*}

\subsection{Result Analysis}
\textbf{Our standalone baseline ($S_1$) results establish that prompt quality is the single largest indicator of standalone agent capabilities.} Deactivating retrieval tools and shifting from full natural language instructions (Descriptive mode) to basic operational declarations (Base mode) incurs a sharp -9pp reduction in accuracy. Stripping the prompt down to vague, localized declarations (Lazy mode) degrades baseline success further by an additional -16pp. 

\textbf{Local workspace environmental feedback via Language Server Protocol (LSP) diagnostics offers negligible compounded mitigation.} Enforcing LSP rules yields a marginal average delta of just +3.7pp across all configurations and helps least where natural language prompt context is adequate, providing a mere +2pp gain in Descriptive mode. This suggests that automated linting feedback does not substitute for the semantic clarity provided by detailed task descriptions.

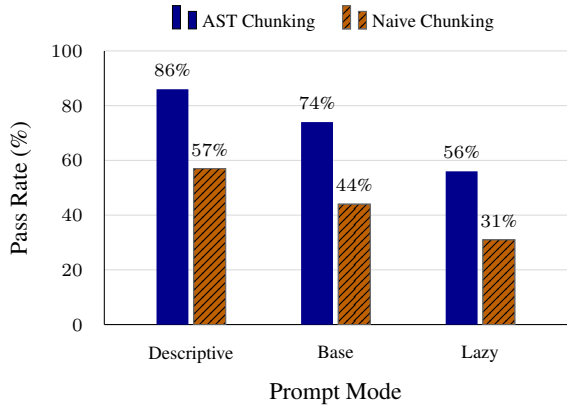
\begin{figure}[htbp]
    \centering
    \begin{tikzpicture}[font=\footnotesize]
    \begin{axis}[
        width=\linewidth, height=5.2cm,
        ybar, bar width=12pt,
        ymin=0, ymax=100,
        ylabel={Pass Rate (\%)},
        xlabel={Prompt Mode},
        label style={font=\footnotesize},
        tick label style={font=\scriptsize},
        symbolic x coords={Descriptive, Base, Lazy},
        xtick=data,
        ytick={0,20,40,60,80,100},
        enlarge x limits=0.3,
        ymajorgrids=true,
        grid style={gray!30, line width=0.3pt},
        axis lines*=left,
        tick align=outside,
        tick style={draw=none},
        nodes near coords={\pgfmathprintnumber{\pgfplotspointmeta}\%},
        every node near coord/.append style={font=\scriptsize, yshift=1pt},
        legend style={
            at={(0.5,1.03)}, anchor=south,
            legend columns=2, draw=none, font=\scriptsize,
            /tikz/every even column/.append style={column sep=8pt}
        },
        legend cell align=left,
    ]
    \addplot[fill=blue!55!black, draw=none] coordinates {(Descriptive,86) (Base,74) (Lazy,56)};
    \addplot[fill=orange!75!black, draw=black!60, postaction={pattern=north east lines}] coordinates {(Descriptive,57) (Base,44) (Lazy,31)};
    \legend{AST Chunking, Naive Chunking}
    \end{axis}
    \end{tikzpicture}
    \caption{Performance comparison between AST Chunking and Naive Chunking across Descriptive, Base, and Lazy prompt modes.}
    \label{fig:s2_compare}
\end{figure}

\textbf{Structural retrieval is the primary reason $S_2$ outperforms $S_1$, and AST-aware chunking is what makes that retrieval effective.} As shown in Table~\ref{tab:main_results} and Figure~\ref{fig:s2_compare}, AST-driven indexing outperforms naive token-window chunking by at least 25pp across every prompt mode. The mechanism is structural: naive methods split code without regard to syntactic boundaries, producing fragments that lack logical scope, whereas AST chunking preserves those boundaries, yielding self-contained functional units that retain the full scope of the retrieved code. This same structural gap drives the broader $S_1$-to-$S_2$ improvement (Table~\ref{tab:final_results}): $S_1$'s \texttt{grep}/\texttt{bash} tooling supports local file navigation but not cross-file reasoning, while $S_2$'s structural context yields an average 9.5pp gain across four model families, confirming that structural awareness, not raw model capability, separates the two regimes.

\textbf{Sub-agent delegation ($S_3$) loses to lean retrieval and does not beat
the baseline.} On the same 100 tasks under Descriptive prompting, $S_2$ scores
86\% against 66\% for $S_3$. The comparison is completely discordant: 20 tasks
pass under retrieval but fail under delegation, with no tasks exhibiting the
reverse outcome. Against $S_1$ the difference is small ($73\%$ vs.\ $66\%$).
Delegation only triggered on 81 of 100 tasks, where the pass rate was 70.4\%;
on the other 19 the sub-agent layer never started and the task ran as a single
agent (47.4\%). We therefore report both figures: 66\% reflects the overall
pass rate when delegation is requested, while 70.4\% reflects performance on
tasks where delegation was actually used. A communication tax
\citep{wang2025agenttaxo} would explain the discordant pattern, but our
telemetry does not separate it from alternatives such as premature scoping by
the parent agent.

\textbf{Retrieval's accuracy gains offset its token overhead, leaving cost per success effectively unchanged.} Structural retrieval increases the raw cost per attempted task (e.g., \$1.07 $\to$ \$1.23 for \texttt{deepseek-v4-pro}), reflecting the additional tokens consumed during indexing and querying. However, Table~\ref{tab:cost_effectiveness_best} shows this overhead is offset by a reduction in wasted attempts: cost per successful refactoring shifts by no more than \$0.01 across all four models evaluated. Structural retrieval's accuracy gains thus come at negligible net cost.

\begin{table}[htbp]
\centering
\small
\setlength{\tabcolsep}{5pt}
\caption{Cost-efficiency at best configuration (Descriptive Mode).}
\label{tab:cost_effectiveness_best}
\begin{tabular}{lccc}
\toprule
\textbf{Configuration} & \textbf{Pass} & \textbf{\$/Task} & \textbf{\$/Pass} \\
\midrule
\texttt{qwen3.6-flash} ($S_1$)   & 73\% & \$0.096 & \textbf{\$0.13} \\
\texttt{qwen3.6-flash} ($S_2$)   & 86\% & \$0.113 & \textbf{\$0.13} \\
\midrule
\texttt{minimax-m3} ($S_1$)      & 75\% & \$0.490 & \$0.65 \\
\texttt{minimax-m3} ($S_2$)      & 81\% & \$0.538 & \$0.66 \\
\midrule
\texttt{kimi-k2.6} ($S_1$)       & 71\% & \$0.444 & \textbf{\$0.63} \\
\texttt{kimi-k2.6} ($S_2$)       & 78\% & \$0.489 & \textbf{\$0.63} \\
\midrule
\texttt{deepseek-v4-pro} ($S_1$) & 77\% & \$1.072 & \textbf{\$1.39} \\
\texttt{deepseek-v4-pro} ($S_2$) & 89\% & \$1.234 & \textbf{\$1.39} \\
\bottomrule
\end{tabular}
\end{table}

\section{Conclusion}
We presented \textbf{RefactorPlatform}, an open-source evaluation harness for repository-scale refactoring agents, providing isolated workspaces, MCP-based retrieval, and auditable per-task logging across model, retrieval, prompting, and orchestration axes. Our demonstration campaign illustrates the platform's core utility: operators can plug in new harnesses and observe their effect on pass rate and cost under controlled conditions. RefactorPlatform is designed to extend to additional benchmarks, verification gates, and orchestration regimes.

\section*{Limitations}
Our current evaluation has three main boundaries driven by practical trade-offs between budget constraints and programming language ecosystems:

\paragraph{Model Scale and Cost} To keep token expenses manageable, we focused on mid-tier models like \texttt{qwen3.6-flash}, \texttt{minimax-m3}, and \texttt{kimi-k2.6}. Full-scale codebase refactoring consumes tokens rapidly, making frontier-scale models costly for large evaluation runs. However, omitting these ultra-large models means we cannot verify if sheer scale naturally fixes the structural blindness or multi-agent penalties we observed. Testing frontier architectures remains an important next step.

\paragraph{Language and Benchmark Scope} Our experiments focus primarily on Python via the RefactorBench dataset. Dynamic languages handle dependencies, imports, and references quite differently than compiled, strongly-typed ecosystems like Java or C++. While we conducted initial trials on the Java-centric \textit{SWE-Refactor} dataset (preliminary results of which are in the appendix~\ref{sec:swe-refactor}) drawing definitive conclusions requires a more dedicated deep dive. Fully mapping our architecture across these distinct language paradigms remains a key focus for our next study.

\paragraph{Campaign scale}
Reported results are single-run campaigns under fixed per-task budgets, so we
report pass rates without uncertainty estimates and describe differences as
observed rather than tested. The same harness supports repeated-trial studies
through configuration and compute alone, without rebuilding isolation,
verification, or telemetry.


\bibliography{custom}
\clearpage

\appendix

\section{SWE-Refactor: Triple-Verification Pipeline}
\label{sec:swe-refactor}
This experiment evaluates product integrity across 1,099 Java refactoring tasks using a stringent multi-gate pipeline that tracks code extraction, compilation success, and structural alignment via \textit{RefactoringMiner}.
To clear the triple-verification pipeline ($S$), a patch must simultaneously extract cleanly, compile, and satisfy \textit{RefactoringMiner} across the evaluation subset (Table~\ref{tab:swe_refactor_results}). The results expose a stark gap between syntax and execution. While single-pass assistants ($S_1$) write highly plausible code: \texttt{minimax-m3} achieves a 97\% extraction rate but the compilation success drops to 34\%, yielding a final task pass rate of just 29\% (51/177 tasks). The core bottleneck is clearly not code generation, but reference propagation across distributed package scopes.

Introducing an execution feedback loop ($S_1$-eval) resolves this drift. By feeding compiler diagnostics back to the model, \texttt{gpt-5-mini}'s compilation rate jumps from 12\% to 88\%, lifting the final pass rate from a 20\% baseline to 85\% (28/33 tasks). In this setting, a compiler-feedback loop outperformed both richer prompting and single-pass generation.

\section{Universal Failure Analysis}

A 5\% subset of tasks fails in all evaluated configurations regardless of model size, prompt engineering, or retrieval architecture. As detailed in Table~\ref{tab:universal_failures}, these persistent failures are not a byproduct of poor context retrieval or weak prompt alignment, but instead point to limitations in multi-file and diff-generation heuristics. For instance, tasks requiring downstream test module updates (\texttt{django}) or deep cross-module parameterization (\texttt{scrapy}) failed uniformly. This indicates that just providing a model the correct code context is insufficient when the underlying change requires complex, multi-step structural edits. Overcoming this 5\% baseline ceiling will require moving past passive retrieval and toward active, execution-guided validation environments.

\section{Detailed Failure Taxonomy}

\begin{table}[htbp]
\centering
\small
\setlength{\tabcolsep}{5pt}
\caption{Universal failure cases.}
\label{tab:universal_failures}
\begin{tabular}{p{2.2cm}lp{2.8cm}} 
\toprule
\textbf{Task Target} & \textbf{Project} & \textbf{Primary Failure Mode} \\
\midrule
\texttt{debug helpers to helpers} & \texttt{flask} & Diff heuristic fails to match changes \\
\midrule
\texttt{new path traversal} & \texttt{django} & Import propagation to test module fails \\
\midrule
\texttt{new utils from basic} & \texttt{ansible} & Incorrect log sanitization relocation \\
\midrule
\texttt{option parser print} & \texttt{tornado} & Outright failure across all 6 variants \\
\midrule
\texttt{parameterize gunzip} & \texttt{scrapy} & Incomplete spider parameterization \\
\bottomrule
\end{tabular}
\end{table}

To identify why automated refactoring degrades at repository scale, we audited 414 failed execution histories across Python and Java testbeds, revealing key model failure boundaries.
The distributions detailed in Table~\ref{tab:combined_failures} confirm that incomplete reference propagation is the primary bottleneck for repository-scale refactoring. On RefactorBench (Panel A), combining the \textit{partial completion} and \textit{missing import} metrics reveals that 50.9\% of failures occur when models apply localized modifications correctly but fail to update dependent call sites. This limitation is more pronounced in SWE-Refactor's compiled Java environment (Panel B), where unresolved package symbols and global import mismatches drive 73.5\% of system failures.

\section{Granular Deep-Dives \& Trace Case Studies}
\label{sec:case_studies}
To anchor our failure taxonomy in concrete real-world contexts, we examine two specific automated execution histories that illustrate the primary breakdown modes observed across our experimental evaluations. As detailed in Table~\ref{tab:case_studies}, these case studies highlight how localized technical successes can ultimately mask systemic architectural blind spots.

\begin{table*}[t]
\centering
\small
\caption{Multi-model empirical evaluation outcomes executed on the SWE-Refactor Java benchmark repository matrix. Due to API rate limits, \texttt{gpt-5-mini} configurations were evaluated on smaller representative subsets ($n=75$ for single-pass baseline; $n=33$ for the feedback loop) compared to the full compound subset ($n=177$).}
\label{tab:swe_refactor_results}
\begin{tabular}{lccccc}
\toprule
\textbf{Model Family Configuration} & \textbf{Setup Regime} & \textbf{Code Extracted} & \textbf{Compiled} & \textbf{AST Verified} & \textbf{Pass Rate} \\
\midrule
\texttt{qwen3.6-flash} & $S_1$ Baseline Default & 90\% & 28\% & 59\% & 24\% (43/177) \\
\texttt{minimax-m3} & $S_1$ Baseline Default & 97\% & 34\% & 55\% & 29\% (51/177) \\
\texttt{gpt-5-mini} & $S_1$ Baseline Default & 82\% & 12\% & 44\% & 20\% (15/75) \\
\texttt{gpt-5-mini} & $S_1$-eval (Feedback Loop) & 94\% & 88\% & 85\% & 85\% (28/33) \\
\midrule
GPT-4o-mini \cite{xu2026swerefactorrepositorylevelbenchmarkrealworld} & Simple Prompting Reference & - & - & - & 19\% (34/177) \\
GPT-4o-mini \cite{xu2026swerefactorrepositorylevelbenchmarkrealworld} & + RAG Few-Shot Baseline & - & - & - & 20\% (36/177) \\
GPT-4o-mini \cite{xu2026swerefactorrepositorylevelbenchmarkrealworld} & Multi-Agent Framework & - & - & - & 66\% (116/177) \\
\bottomrule
\end{tabular}
\end{table*}

\begin{table*}[t]
\centering
\small
\caption{Granular failure taxonomy across datasets, detailing the primary breakdown categories on RefactorBench (Panel A) and the compilation pipeline execution failures on SWE-Refactor (Panel B).}
\begin{tabularx}{\textwidth}{Xcc}
\toprule
\textbf{Failure Category / Pipeline Stage} & \textbf{Observed Failures} & \textbf{\% of Total Failures} \\
\midrule
\multicolumn{3}{l}{\textbf{Panel A: RefactorBench (Python Dataset, 300 exploratory agent runs)}} \\
\midrule
Partial completion (some subtests pass, missed 1--2 trailing files) & 31 & 27.7\% \\
Assertion / functional logic error (incorrect refactored behavior) & 30 & 26.8\% \\
Missing import statement / broken variable reference post-move & 26 & 23.2\% \\
All subtests fail (complete task miss) & 13 & 11.6\% \\
Harness execution timeout (agent caught in tool-invocation loops) & 10 & 8.9\% \\
No structural repository edits produced & 2 & 1.8\% \\
\cmidrule(lr){1-3}
\textbf{RefactorBench Total} & \textbf{112} & \textbf{100.0\%} \\
\midrule
\multicolumn{3}{l}{\textbf{Panel B: SWE-Refactor (Java Dataset Pipeline Stages)}} \\
\midrule
Compilation failed (missing package symbols/unresolved class paths) & 222 & 73.5\% \\
AST verification failed (structural transformation layout type mismatch) & 80 & 26.5\% \\
\cmidrule(lr){1-3}
\textbf{SWE-Refactor Total} & \textbf{302} & \textbf{100.0\%} \\
\bottomrule
\end{tabularx}
\label{tab:combined_failures}
\end{table*}

\begin{table}[htbp]
\centering
\small
\caption{Case Studies of Agent Architectural Failures}
\label{tab:case_studies}
\begin{tabularx}{\textwidth}{l p{3.2cm} X X}
\toprule
\textbf{Repository} & \textbf{Objective \& Success} & \textbf{Blind Spot \& Breaking Point} & \textbf{Architectural Diagnosis} \\
\midrule
\textbf{Ansible} & 
\textbf{Obj:} Extract core utility functions to a new \texttt{utils.py}. \newline 
\textbf{Succ:} Successfully refactored 9 production import sites. & 
Treated production and test suites as isolated; failed to audit test files for stale assignments. \newline 
\textbf{Fail:} 2/12 tests failed (\texttt{AssertionError} in test suite). & 
\textbf{Partial Completion.} Handles immediate local edits successfully, but lacks the repo-wide scope to synchronize downstream testing environments. \\
\midrule
\textbf{Django} & 
\textbf{Obj:} Add optional \texttt{log=True} flag to \texttt{get\_resolver()} and propagate. \newline 
\textbf{Succ:} Core definition block modified accurately. & 
Assumed strictly localized scope; skipped repo-wide search, missing 2 critical caller nodes. \newline 
\textbf{Fail:} 2/3 tests failed (\texttt{AssertionError}: flag unexpectedly \texttt{None}). & 
\textbf{Reference Blindness.} Local technical edits succeed, but failure to perform codebase-wide discovery leaves external dependencies broken. \\
\bottomrule
\end{tabularx}
\end{table}

\end{document}